\documentclass{article}

      \usepackage[final]{neurips_2019}

\usepackage[utf8]{inputenc} % allow utf-8 input
\usepackage[T1]{fontenc}    % use 8-bit T1 fonts
\usepackage{hyperref}       % hyperlinks
\usepackage{url}            % simple URL typesetting
\usepackage{booktabs}       % professional-quality tables
\usepackage{amsfonts}       % blackboard math symbols
\usepackage{nicefrac}       % compact symbols for 1/2, etc.
\usepackage{microtype}      % microtypography
\usepackage[ruled,vlined]{algorithm2e}
\usepackage{graphicx}
\usepackage{float}
\usepackage{appendix}
\usepackage{subcaption}
\title{Automated curriculum generation for Policy Gradients from Demonstrations}
\author{%
  Anirudh Srinivasan \\
  Microsoft Research\thanks{Work done during an internship at Mila}\\
  \texttt{anirudhsriniv@gmail.com} \\
  \And
  Dzmitry Bahdanau \\
  Mila, Université de Montréal \\
  \AND
  Maxime Chevalier-Boisvert \\
  Mila \\
  \And
  Yoshua Bengio \\
  Mila, Université de Montréal \\
}

\begin{document}

\maketitle

\begin{abstract}
  In this paper, we present a technique that improves the process of training an agent (using RL)
 for instruction following.
  We develop a training curriculum that uses a nominal number of expert demonstrations and trains the agent in a manner that draws parallels from one of the ways in which humans learn to perform complex tasks, i.e by starting from the goal and working backwards.
  We test our method on the BabyAI platform and show an improvement in sample efficiency for  some of its tasks compared to a PPO (proximal policy optimization) baseline.
  % We show that some of the levels show improvements in sample efficiency compared to a PPO (proximal policy optimization) baseline.
  % In this paper, we show that by using a small number of expert demonstrations,
\end{abstract}

\section{Introduction}
Training an agent for understanding and following human language instructions to complete a task is an active area of research. The end goal is to have an agent that is able to take in a task in human language and execute it in the environment.
More recently, several such environments have been developed that are easy to use and experiment on by an end user (\citet{hermann2017grounded}, \citet{yu2018interactive} and \citet{chevalier-boisvert2018babyai}).

One such platform is BabyAI \citet{chevalier-boisvert2018babyai}, which is presented as a tool to study the sample efficiency of different algorithms for this task. It comprises of multiple levels, with each level having a gridworld environment, and an agent, that is given a task in natural language that it has to accomplish.
Each level is comprised of different challenges (like navigation, mazes, distractors, door unlocking, sequencing of tasks etc.) that the neural agent has to learn.
% levels with varying difficulty.  has to understand and learn. They present their platform as a means of training an agent with a human input in the loop. % They demonstrate the use of curriculum learning in their scenario and talk about how it is an essential aspect to training with a human in the loop.

Research in this area has moved on from just training an agent to complete a task. Nowadays, the focus is a lot on visualizing and understanding how the agent learns. One of the areas being explored is training the agent in a way that is similar to how humans learn a particular task. The best example of this is curriculum learning, i.e training a model for a simpler task first so that it can be trained on a complex task more easily.
\citet{Bengio:2009:CL:1553374.1553380} and many others have shown that this does work for neural networks.
Another of the methods that humans use to solve complex tasks is to start off from the goal and work backwards. \citet{pmlr-v78-florensa17a} propose a technique where a robots can be trained on a curriculum based on this idea.

This paper proposes a algorithm that is inspired by this method that humans learn complex tasks.
Our method comprises of a curriculum, that trains the agent on tasks right next to the goal first, introducing tasks that are increasingly more difficult.
To obtain these tasks, we use information from expert demonstrations.
Our method shows improvements on BabyAI levels whose demonstrations are not extremely long and require a decent amount of exploration by the agent to solve.
% and report on when f exploration our technique works and when it doesn't.

\section{Related Work}
One category of methods for instruction following is imitation learning, where a neural network maps each input to the action to execute.
This category ranges from simple methods like behavioral cloning to more complex ones like DAGGER \citet{ross2010reduction} that have a human involved in the process.
Another family of methods are reinforcement learning based, where the agent has to learn a policy \citet{sutton2000policy} or a Q-function \citet{watkins1992q}.
A final class of methods which do not directly model the policy exist, under which methods like maximum entropy inverse reinforcement learning \citet{ziebart2008maximum} and GAIL \cite{ho2016generative} fall in.

\citet{Bengio:2009:CL:1553374.1553380} proposed curriculum learning and showed that pre-training an LSTM (for language modelling) with easier tasks meant training on more complex tasks could happen much faster. 
There have been multiple attempts of applying curriculum learning to a reinforcement learning task. Most of these methods are concerned with determining the order to present the tasks of the curriculum.
% \citet{svetlik2017automatic} propose a method where task descriptors are used to order tasks in the form of a directed acyclic graph, thus obtaining a curriculum.
% \citet{silva2018object} assign an object oriented representation to each task so that certain learned knowledge can be reused.
Methods have been proposed where descriptions of the task are used to construct a directed acyclic graph of tasks (\citet{svetlik2017automatic}) or an object oriented representation of the tasks (\citet{silva2018object}), from which the curriculum is obtained.
\citet{matiisen2017teacher} (Teacher-Student curriculum learning), \citet{graves2017automated} and \citet{narvekar2017autonomous} (Curriculum policies) all propose methods where the task of selecting the next task to train on is viewed as an RL problem in itself.
Most of these methods rely on a curriculum already having been developed by the user.
\citet{pmlr-v78-florensa17a} propose a method where the curriculum is formulated by exploring the states of the environment and ordering them.

Some works have also proposed the use expert demonstrations during the reinforcement learning training process.
\citet{8463162} used it to speedup the training process on robotics tasks with sparse reward functions. They store the demonstrations in an additional replay buffer from which they sample from during a minibatch. They also use the states from the demonstrations to reset the agent to. Both these speed up the initial phase of the training process, where the agent ends up getting a zero reward quite frequently.
The work of \citet{DBLP:journals/corr/HesterVPLSPSDOA17} also use demonstrations via the form of a replay buffer that they sample from.
Finally, the work of \citet{resnick2018backplay} propose a technique called Backplay where they use a single demonstration
to create a curriculum for the agent.

\section{Reverse Curriculum Learning}
\subsection{Existing Work}
\citet{pmlr-v78-florensa17a} propose Reverse Curriculum generation. In this method,
the agent starts off from a state right next to the goal state and begins a random walk.
The states that are at one step away form the first stage of the curriculum, two steps
form the second stage and so on. In this way, a curriculum is built without needing
any intervention from the user. Their algorithm is described in \ref{alg: rcl}.

\begin{algorithm}
  \DontPrintSemicolon
  \KwData{$goal\_state$: final state of agent, $n\_stages$: number of stages in curriculum wanted}
  \KwResult{$curriculum$: list of stages, with each stage as a list of start states}
  $curriculum$ $\gets$ [ [ ] repeated $n\_stages$ times]\;
  $states$ $\gets$ [ $goal\_state$ repeated $n$ times]\;
  \For{i = 1 to n\_stages}{
      $states_{new}$ $\gets$ [ ]\;
      \For{state in states}{
          Sample random $action$\;
          $new\_state$ $\gets$ $state$.step($action$)\;
          $states_{new}$.append($new\_state$)\;
      }
      $curriculum$[i] $\gets$ $stages_{new}$\;
      $states$ $\gets$ $states_{new}$\;
  }
  \caption{Reverse Curriculum Learning}
  \label{alg: rcl}
\end{algorithm}

\subsection{Applying existing methods to our environment}
In \citet{pmlr-v78-florensa17a}'s work, the above mentioned technique achieved good results on robotic arm movement type tasks.
These tasks are different from our case as they a state and action space that is continuous. BabyAI's gridworld environment has a state and action space that is discrete.
We observed that if an agent starts a random walk form near the goal, it ends up at the goal state quite frequently (see Table \ref{tab:pergoal}), rendering that path generated useless.
This exploration technique is not the right one to use in the case of a discrete environment.

\begin{table}[h]
\centering
\begin{tabular}{llllll}
\toprule
\textbf{Env/Steps}    & \textbf{1}    & \textbf{2}    & \textbf{3}   & \textbf{4}    & \textbf{5}    \\
\midrule
GoToLocal    & 34.4 & 10.2 & 4.5 & 3.6  & 3.1  \\
PutNextLocal & 26.5 & 6.3  & 1.8 & 0.01 & 0.01 \\
\bottomrule
\end{tabular}
\caption{\% of times the agent reaches the goal when the it starts 1,2,3,4 and 5 steps away and performs a random walk}
\label{tab:pergoal}
\end{table}
% Table on number of times

\subsection{Using demos to generate a curriculum}
\begin{figure}
  \centering
  \includegraphics[width=0.8\textwidth]{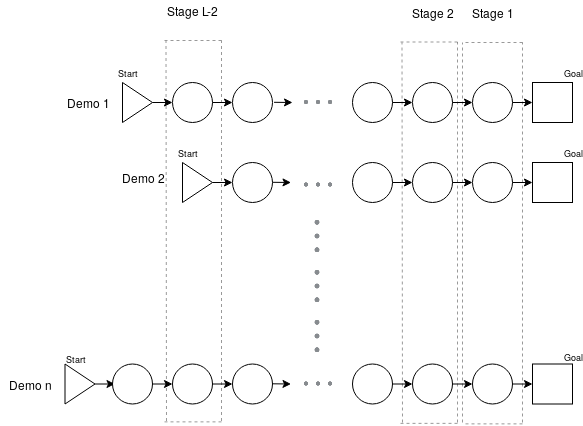}
  \caption{Generation of curriculum stages from demos. Triangles are \textit{start} states, Squares 
  are \textit{goal} states and Circles are \textit{intermediate} states}
  \label{fig:rcppo}
\end{figure}

To alleviate this, we come up with a method that uses the demonstrations generated by BabyAI's heuristic expert.
Each demonstration comprises of a set of actions to perform from a start state to reach the end/goal state.
Traversing through each demo, we obtain the states that are one step/action away from the
goal state and make this the first stage of the curriculum. We repeat this process and obtain steps
that are 2 steps away from the goal state and make this the second stage of the curriculum. This
process is repeated to obtain the entire curriculum. The algorithm is detailed in \ref{alg: currdemo}.
Figure \ref{fig:rcppo} has a graphical depiction of how the curriculum is generated.

\begin{algorithm}
  \DontPrintSemicolon
  \KwData{$demos$: array of demos generated by the bot}
  \KwResult{$curriculum$: list of stages, with each stage as a list of start states}
  $curriculum\_length$ $\gets$ max([len($demo$) for $demo$ in $demos$]) - 1\;
  $n\_demos$ $\gets$ len($demos$)\;
  $curriculum$ $\gets$ [ [ ] repeated $curriculum\_length$ times]\;
  \For{demo in demos}{
    $n\_steps$ $\gets$ len($demo$) - 1\;
    \For{i = 1 to n\_steps}{
      Initialize $env$ with seed used for $demo$\;
      Step through $env$ $i$ times based on $demo$\;
      add a copy of $env$ to $curriculum[n\_steps - i - 1]$\;
      }
      }
      \caption{Generating curriculum from demos}
      \label{alg: currdemo}
    \end{algorithm}

The curriculum is defined as an ordered set of stages, with each stage being a set of start states as obtained by Algorithm \ref{alg: currdemo}.
The number of stages in the curriculum(i.e curriculum length) is determined by the maximum length of the demos given.
This technique can be applied to any problem being solved by RL where a small number of demonstrations are available.
The ordering of tasks in this method is similar to how humans learn to perform complex tasks, by starting off from something closer to the goal and working backwards.

We call our modified algorithm RCPPO, as it builds up on PPO \citet{schulman2017proximal}.
PPO is modified in such a way that each time the agent resets after the end of an episode, it is allowed to only reset to the set of states allowed as per the stage of the curriculum it is in.
This is very similar to \citet{8463162}, where the agent may chose to reset to any random state or a state from the demonstration.
Once all the stages in the curriculum have been completed, the agent is allowed to reset to any state in the environment.

\section{Experiments}
We compared our algorithm against an implementation of PPO, which has built been
built up on Policy Gradients \cite{sutton2000policy}. We tested our algorithm on the BabyAI as it has a large number of levels, each level composed of different challenges that the network has to learn.
For each level, we report the number of frames needed to be seen before the agent reaches a particular accuracy (0.95/0.99) on that level.
For methods that use the curriculum, we check for accuracy only after the curriculum has been completed (i.e on the task where it is allowed to reset to any state on the grid).
Success in BabyAI is defined as the agent getting a reward greater than 0, which happens as long as it reaches the goal state.

We ran 2 versions of RCPPO, with the difference between them being in the criterion used to determine when to move to the next stage of the curriculum.
The first variant would make a curriculum update when the success for that stage hit 90\% percent. The second variant would set this threshold to 70\% and gradually increase it to 99\% as the stages in the curriculum pass through.
We report the mean of the two methods in the results. 
We evaluated other methods for this, methods like \cite{matiisen2017teacher} (refer to \ref{app:tc} in Appendix), but determined that a simple method like this was enough.

In our experiments, we used 1000 demonstrations from the expert for all the levels. We also performed a study on how many demonstrations are needed to do well enough (see \ref{app:demo} in Appendix), however that is not the main focus of our work.

Code containing implementation of the above technique is available at \footnote{Code: \url{https://github.com/Genius1237/babyai/tree/rcppo}}

\begin{table}[h]
  \centering
  \begin{tabular}{lll}
    \toprule
    \textbf{Level}               & \textbf{PPO}    & \textbf{RCPPO}  \\
  \midrule
  \textit{Small Levels} & \textit{(For 0.99} & \textit{Accuracy)} \\
  % GoToRedBallGrey    & 9680 & 10240 \\
  % UnlockLocal     & 6144 & 11264 \\
  GoToRedBall         & 34816  &  41728      \\
  PickupLoc         & 95488 & 205056 \\
  GoToObjMaze       & 134144 & 176128 \\
  GoToLocal           & 138240 &  140288      \\
  PickupLocal         & 177920 & 205056       \\
  PutNextLocal        & 804096 & \textbf{580608} \\
  UnlockPickup        & -      &  \textbf{15360}      \\
  BlockedUnlockPickup & -      &  \textbf{26112}      \\
  UnlockPickupDist    & -      &  \textbf{366080}      \\
 
  \midrule
  \textit{Large Levels} & \textit{(For 0.95} & \textit{Accuracy)} \\
  Open                & 72960 & 101632      \\
  GoTo                &  577536      & 787712              \\
  Pickup            & 762368 & - \\
  \bottomrule
  \end{tabular}
  \caption{\# Frames (in100s) to reach that accuracy. "-" indicates level was not solvable}
\end{table}

\section{Results and Analysis}
% Based on the mean demo length (refer to \ref{tbl:demol} in Appendix), 
We divided the levels into 2 categories: small and large, based on the number of rooms in the environment.
We report our findings for both separately

\subsection{RCPPO on small levels}
\begin{figure}[h]
  \centering
  \begin{subfigure}[b]{0.4\textwidth}
      \centering
      \includegraphics[width=\textwidth]{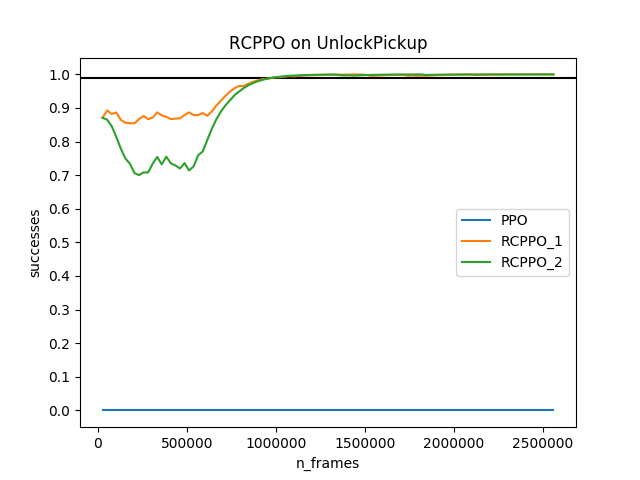}
  \end{subfigure}    
  \begin{subfigure}[b]{0.4\textwidth}  
      \centering 
      \includegraphics[width=\textwidth]{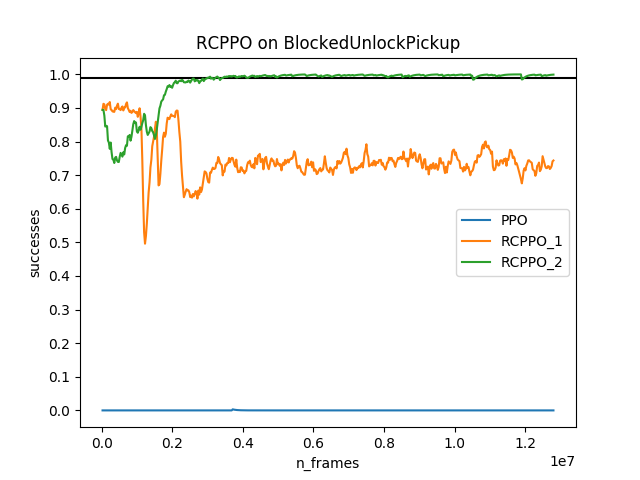}
  \end{subfigure}    
  \begin{subfigure}[b]{0.4\textwidth}   
      \centering 
      \includegraphics[width=\textwidth]{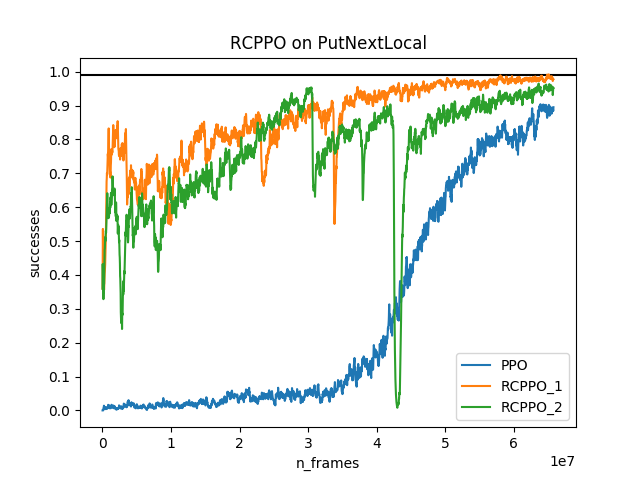}
  \end{subfigure}    
  \begin{subfigure}[b]{0.4\textwidth}   
      \centering 
      \includegraphics[width=\textwidth]{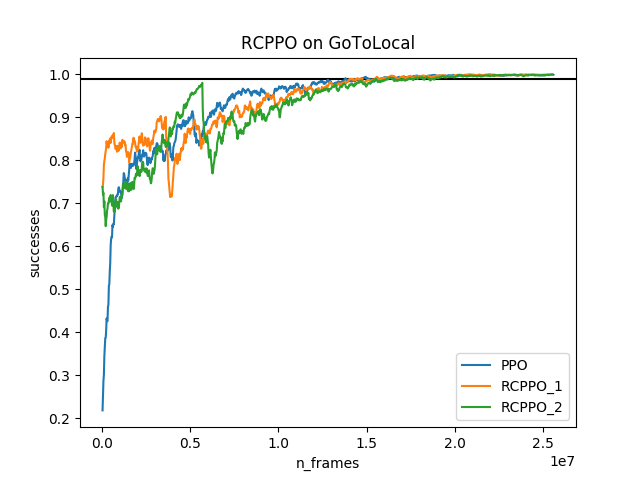}
  \end{subfigure}    
  \caption[]{\small RCPPO on small levels} 
  %\label{fig:mean and std of nets}
\end{figure}

Small levels were levels with 1 room(GoToLocal, GoToRedBall etc..) or 2 rooms(UnlockPickup, BlockedUnlockPickup). For these levels, we report the number of frames to reach 0.99 accuracy, as that is easily achievable by our baseline.

\begin{enumerate}
  \item Levels like GoToRedBall, GoToLocal etc.., show no improvement by using
        our technique.
  \item Levels like PutNextLocal, UnlockPickup etc.. show significant improvements by using our method.
        In fact, levels like UnlockPickup, UnlockPickupDist and BlockedUnlockPickup are unsolvable
        by vanilla PPO but our method is able to solve them extremely easily.
\end{enumerate}

Levels like GoToRedBall, GoToLocal are extremely simple. These require the agent to only perform 2 or 3 (movement related) of it's 7 possible actions to solve each level. We theorize that a neural network is easily able to learn this and hence our method is ineffective.

On the more complex levels, the agent has to execute more of the actions from it's space of possible actions to solve each level.
The mean demo length for these levels also ends up being larger (refer to Table \ref{tbl:demol} in Appendix).
Our method is also able to handle the problem of having a lot of zero reward episodes in the initial training stages (as described in \citet{8463162}).
Our curriculum is able to help the agent handle all of these and speed up it's training.

\subsection{RCPPO on large levels}
\begin{figure}[h]
  \centering
  \begin{subfigure}[b]{0.4\textwidth}
      \centering
      \includegraphics[width=\textwidth]{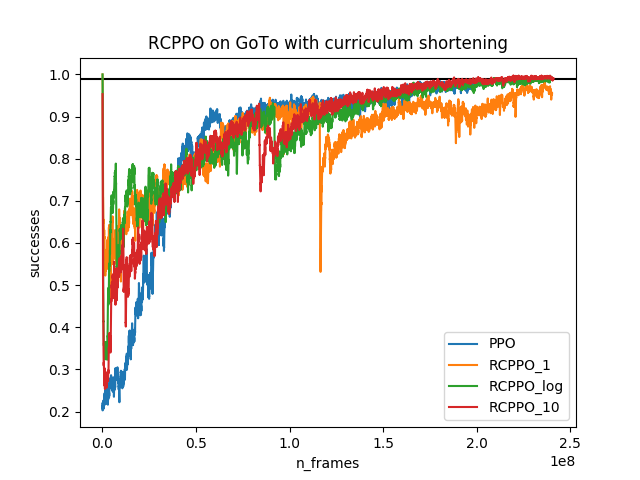}
      \caption{GoTo}
      \label{fig:GoTo}
  \end{subfigure}    
  \begin{subfigure}[b]{0.4\textwidth}  
     \centering 
     \includegraphics[width=\textwidth]{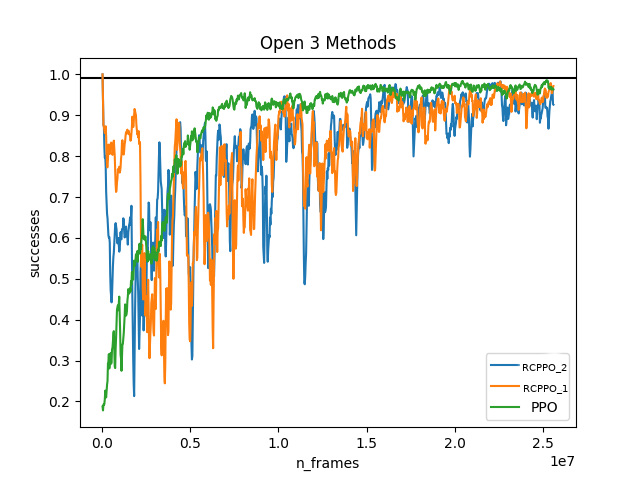}
     \caption{Open}
  \end{subfigure}
  \caption{RCPPO on large levels, with variations}
\end{figure}

Large levels were levels with more than 2 rooms. These levels are characterized by extremely long
demos produced by the bot (\ref{tbl:demol}). For these levels, we report the number of frames to reach 0.95 accuracy, as reaching 0.99 was difficult, even for the baseline method.
For our analysis, we look at the performance of GoTo, as the performance
for other levels were similar to it.

In the case of GoTo (RCPPO\_1 in \ref{fig:GoTo}), the curriculum ends at an extremely late stage, due to which the technique does
not show result in an improvement. This is very much akin to a statement made in
\cite{Bengio:2009:CL:1553374.1553380}, where they say that using curriculum
learning will result in the model needing to see more examples while training.

To combat this problem, we implemented a method where we combined $n$ consecutive stages down to one stage, to reduce the curriculum length.
We tried multiple methods, with combining 5 stages into one, 10 stages into one and finally in an exponential sense (1, 2, 4.... so
on). As is evident from \ref{fig:GoTo}, this method also did not result in much improvement.
We conclude that even after combining stages together, the tasks within each stage are too varied and the agent isn’t able to learn effectively.

Similar performance is observed for other large levels. We conclude that our method is not effective when the
mean demo length is very large.

\section{Conclusions}
We present our algorithm RCPPO, that takes in some demonstrations from the an expert, and uses it to build a curriculum to train an agent for
instruction following. We observe improvements on levels where the agent has to execute a wide variety of actions to reach the goal, as long as
the number of actions to reach the goal is not too large.

Our work builds up on top of the work by \citet{pmlr-v78-florensa17a}. Their method needed a bit of tweaking to work in the action and state spaces present in our case. Rather than using a random walk to expand the state space for further stages of the curriculum, we use the information form the demos and constrain the expansion to the states in the demos. We end up getting a fixed length curriculum for our task.

In our work, we were not able to evaluate our technique on the most difficult of levels in BabyAI. Although our experiments with GoTo suggest that
it is unlikely to work on more complex levels, a more refined version of our method
could be tried out on the larger levels like SeqToSeq and PutNext. One could look at a particular start state in the curriculum and use the series of observations that were obtained while traversing through the demonstration to obtain that start state and use this information in a recurrence, similar to how memory is incorporated into policy gradients by \cite{wierstra2007solving}. This information may help is solving levels that have long demonstrations.

Although a short study on it is presented in the appendix \ref{app:demo}, we have not done extensive analysis on how the number of demos used affects learning. This is also something that could be looked into.

\subsubsection*{Acknowledgments}
We would like to thank Léonard Boussioux and David Yu-Tung Hui for their
inputs and motivation during the experimentation for this work.
We would like to thank Sebastin Santy for his feedback on the draft of the paper.
We would also like to thank Compute Canada for providing the GPU resources used to
run the experiments for this work.

\small
% \nocite{*} % Insert publications even if they are not cited in the poster
\bibliography{ref}
\normalsize
% \appendix
\section*{Appendix}
% \appendix
\subsection*{Statistics on demonstrations}
We computed the mean and max length of demos for different levels. The max length of the demos corresponds to the length of the curriculum for that level.
\begin{table}[h]
  
  \centering
  \begin{tabular}{lll}
    \toprule
    \textbf{Level}               & \textbf{Avg}    & \textbf{Max}  \\
    \midrule
    GoToRedBall         & 6.2 & 21        \\
    GoToLocal           & 6.4 & 23      \\
    GoToRedBallGrey & 6.8 & 24 \\
    PickupLocal         & 7.0 & 23       \\
    PutNextLocal        & 13.0 & 56 \\
    UnlockLocal & 15.1 & 28 \\
    UnlockPickup        & 20.4 & 31      \\
    UnlockPickupDist    & 26.7 & 68      \\
    Open                & 30.1 & 186      \\
    BlockedUnlockPickup & 35.3 & 47      \\
    GoTo & 52.9 & 208 \\
    Pickup & 53.9 & 209\\
    GoToSeq & 72.2 & 265 \\
    PutNext & 90.2 & 237 \\
    \bottomrule
\end{tabular}
\caption{Avg and Max length of demos. Determines curriculum lengths}
\label{tbl:demol}  
\end{table}

\subsection*{Effect of number of demos used to build curriculum}
\label{app:demo}
We perform a study to see how the performance of training the agent changes when different number of demos are used to build a curriculum. To this end, we perform training on PutNextLocal with 100,500,1000 and 2000 demos. We report the same metric as before, number of frames to reach a particular success level, and this time, we include that statistic for success rates of 0.9 and 0.95 as well. As is evident, the number of demos used does not seem to have a significant impact on the training, with all our results being close to each other.

\begin{minipage}[b]{0.49\textwidth}
 \centering
% \begin{table}[b]
\begin{tabular}[b]{llll}
\toprule
\textbf{\#Demos} & \textbf{0.9}    & \textbf{0.95}   & \textbf{0.99}   \\
\midrule
% PPO            & 637440 & 688384 & 804096 \\
100            & 300800 & 348672 & 584448 \\
500            & 348160 & 502016 & 655616 \\
1000           & 288256 & 371712 & 580608 \\
2000           & 374016 & 479232 & 611840 \\
\bottomrule
\end{tabular}
\captionof{table}{\#Frames to reach particular success level}
\label{tab:framesucc}
% \end{table}
\end{minipage}
\hfill
\begin{minipage}[b]{0.49\textwidth}
    \centering
    \includegraphics[width=0.99\textwidth]{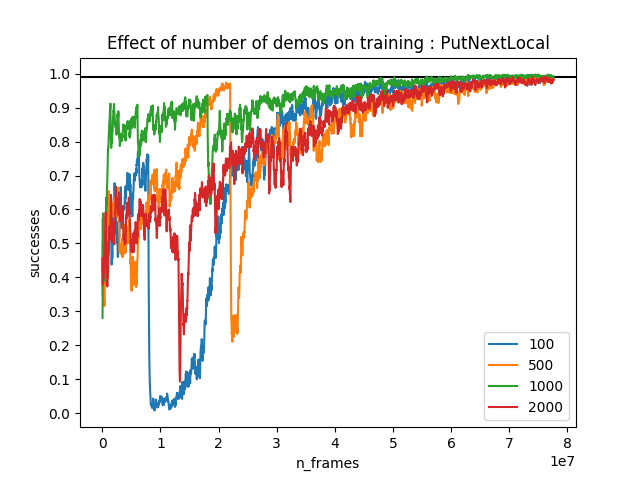}
    \captionof{figure}{Training curves using different \#demos}
    \label{fig:my_label}
\end{minipage}

\subsection*{Teacher-Student Curriculum Learning}
\label{app:tc}
\begin{figure}[h]
    \centering
    \includegraphics[width=0.5\textwidth]{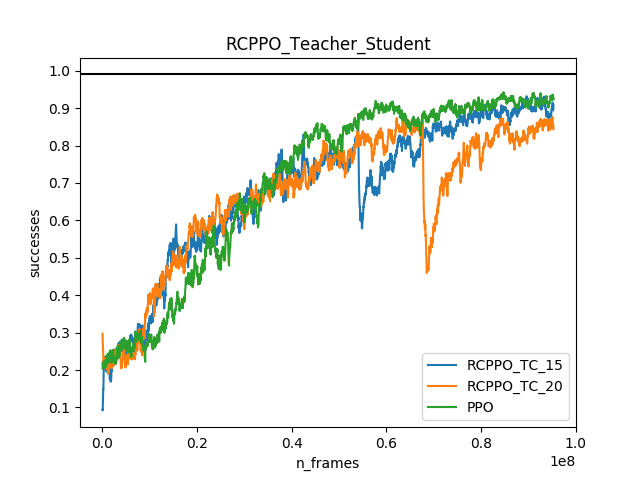}
    \caption{Teacher Student curriculum learning on RCPPO vs vanilla PPO}
    \label{fig:tcrcppo}
\end{figure}

We used teacher student curriculum learning (\cite{matiisen2017teacher}) to determine when/what curriculum stage to change to. We evaluated this for GoTo, the level on which our algorithm was struggling. We did not see any improvement. Training curves are shown in \ref{fig:tcrcppo} depict 2 different versions of teacher student curriculum against PPO.
\end{document}